\documentclass[runningheads]{llncs}
\usepackage[T1]{fontenc}
\usepackage{graphicx}
\usepackage{booktabs} 
\usepackage{array}
\usepackage{multirow}
\usepackage{color}
\usepackage{bbding}
\begin{document}
\title{Contrastive Centroid Supervision Alleviates Domain Shift in Medical Image Classification}
%
%
\author{Wenshuo Zhou\inst{*} \and
Dalu Yang\inst{*} \and
Binghong Wu \and
Yehui Yang\dag \and
Junde Wu \and
Xiaorong Wang \and
Lei Wang \and
Haifeng Huang \and
Yanwu Xu}

\institute{Intelligent Healthcare Unit, Baidu Inc, Beijing, China \\
\inst{*} equal contribution \\
\dag corresponding author and project leader}

\maketitle              

\begin{abstract}
Deep learning based medical imaging classification models usually suffer from the domain shift problem, where the classification performance drops when training data and real world data differ in imaging equipment manufacturer, image acquisition protocol, patient populations, etc.
We propose Feature Centroid Contrast Learning (FCCL), which can improve target domain classification performance by extra supervision during training with contrastive loss between instance and class centroid. 
Compared with current unsupervised domain adaptation and domain generalization methods, FCCL performs better while only requires labeled image data from a single source domain and no target domain.
We verify through extensive experiments that FCCL can achieve superior performance on at least three imaging modalities, i.e. fundus photographs, dermatoscopic images, and H\&E tissue images.


\end{abstract}
\section{Introduction}
Deep learning models have been widely used in medical image classification tasks\cite{ref_article0,ref_article11}.
However, these models suffer from the domain shift~\cite{ref_book1} problem caused by various factors including imaging equipment from different manufacturers, different data acquisition protocols, different patient populations, etc. 
When a model trained source domain model is applied to the target domain, the performance may drop significantly. 
Therefore, researches on the domain shift problem have emerged and attracted increasing interest over recent years\cite{ref_article3,ref_article9}. 

Among all the efforts to overcome the domain shift problem, Unsupervised Domain Adaptation (UDA)\cite{ref_article2,ref_article8} and Domain Generalization (DG)\cite{ref_article10} methods are the most commonly adopted ones in the field of medical image classification tasks\cite{ref_article6}. 
This is because UDA and DG methods can work when target domain images and/or labels are hard to acquire for training, which is a fairly common situation in real world medical studies. 

However, current UDA and DG researches still have two weaknesses: 
1) UDA and DG only work when images from at least two domains are available, because most of these methods depend on domain discrepancy~\cite{ref_proc1,ref_article14,ref_proc2,ref_proc3,ref_proc4,ref_proc5} or domain adversarial~\cite{ref_article4,ref_proc11,ref_proc12} training. 
UDA in general requires labeled images from a single source domain and unlabeled images from target domains, while DG requires labeled images in multiple source domains. 
This can be a problem when only labeled images from a single source domain is available, e.g. medical imaging manufacturer developing deep learning models for a new imaging device when the only available data are from an older device model. 
2) Most UDA and DG researches in medical image classification only verify their algorithm on a single modality, thus the generalizability of the adaptation methods remains to be investigated. 



In this paper, we propose a new Feature Centroid Contrast Learning (FCCL).
Inspired by recent contrastive learning researches~\cite{ref_proc16,ref_proc15}, 
our algorithm can mitigate the domain-shift-induced performance degradation by increasing/decreasing the inter/intra-class distance of features. 
The main contributions of this paper are: 

1) FCCL only requires labeled image data from a single source domain, while performs better than most UDA methods that requires unlabeled target domain images. 

2) FCCL can achieve better performance on target domain images with at least three imaging modalities, namely fundus photographs, dermatoscopic images, and H\&E tissue images. 

\begin{figure}
\includegraphics[width=\textwidth]{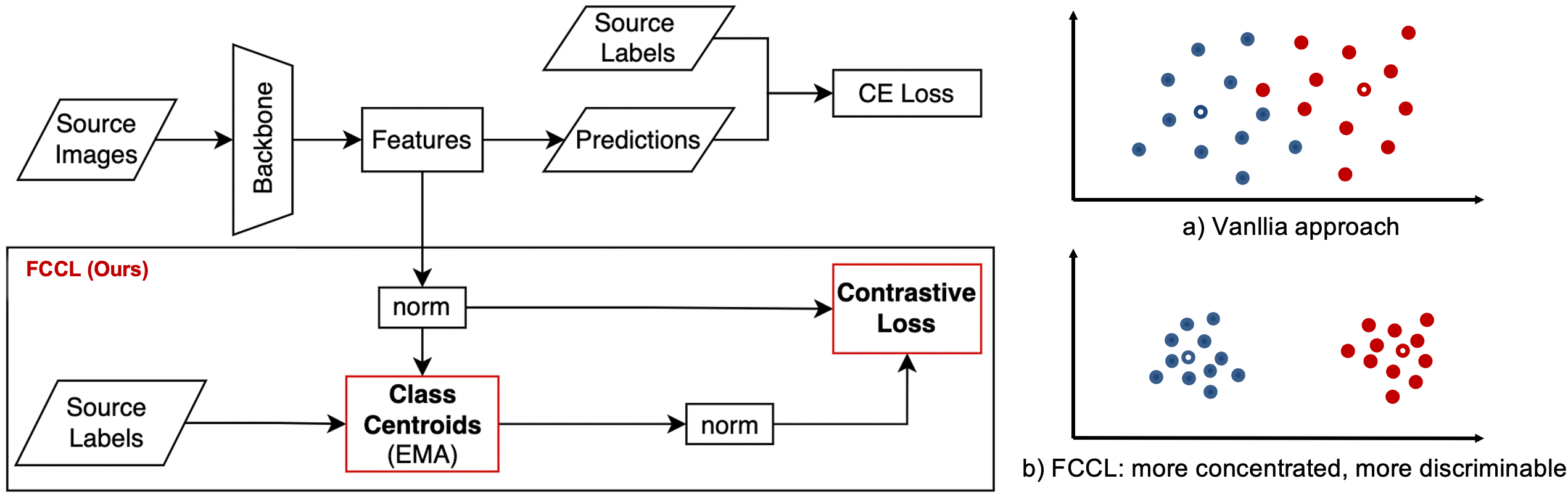}
\caption{Feature Centroid Contrast Learning. Our contrastive loss module works directly on the features of the backbone output, increasing interpretability, as shown on the right. Different from other contrastive learning, we only use one data augmentation and one backbone. The \textbf{norm} module represents $l_2$ normalization, which ensures stable changes in the centroid.}
\label{FCCL}
\end{figure}

\section{Method}
We propose Feature Centroid Contrast Learning (FCCL), as shown in Fig~\ref{FCCL}. When training on the source domain, the model is constrained by a contrastive loss (Fig.~\ref{FCCL} bottom branch) in addition to the regular cross-entropy loss (Fig.~\ref{FCCL} top branch). 
The contrastive loss is computed between the features generated by the backbone of the classifier network and the class centroids of these features. 
Intuitively, this loss will pull each sample towards the class centroid it belongs to, and push the sample away from all the centroids of other classes. See Fig.~\ref{FCCL} on the right for a visualization of how the features of each instance move with the centroids. 





FCCL constrains the distribution of different classes of data features through feature and centroids Contrast. The contrast loss is:
\begin{equation}
\mathcal{L}_{i}^{cont} =-\log{}{\frac{\exp\left ( f_{i} \cdot c^{+}/\tau \right ) }{\exp\left ( f_{i} \cdot c^{+}/\tau \right ) +\sum_{c^{-} }^{} \exp\left ( f_{i} \cdot c^{-}/\tau \right ) } }.
\end{equation}
Here, $f_i$ represents the feature vector output by the backbone corresponding to sample $i$. $c^+$ represents the class centroid of the class in which the sample $i$ belongs, as the positive sample corresponding to the sample $i$. The set $\{c^-\}$ corresponds to all the class centroids of different classes from sample $i$, which are used as negative samples of sample $i$. The hyperparameter $\tau$ is the temperature coefficient to adjust the distribution of $f_i$ and $c$.

In order to efficiently obtain the class centroids of the entire dataset, the class centroids are updated by means of EMA. The formula is:

\begin{equation}
c{\leftarrow} mc + \left ( 1 -m\right ) f_{mean}.
\end{equation}
Among them, $f_{mean}$ is the class center calculated in the current minibatch, which is obtained by averaging the features of each class of samples. $m$ represents the smoothing coefficient, in order to stabilize the centroid, the update formula is:
\begin{equation}
m{\leftarrow} m_0 + \left ( 1 -m_0\right ) \frac{e}{A} .
\end{equation}
Among them, $m_0$ represents the initial value, $e$ represents the current epoch number, and $A$ represents the total number of epochs.
The classifier is a single layer fully connected layer, using the cross-entropy loss function, the formula is:

\begin{equation}
\mathcal{L}_{i}^{ce} =- {\sum_{j}^{} }{} y_{ij}\cdot \log_{}p_{ij}.
\end{equation}
Here, $y_{ij}$ is the label corresponding to sample $i$, and $p_{ij}$ is the probability value of the softmax output of sample $i$.

The loss function of our proposed FCCL model includes classification loss and contrast loss, and the final loss function is shown in formula (4), where $\alpha$ is the weighting coefficient of contrast loss.

\begin{equation}
\mathcal{L}=\mathcal{L}^{ce} + \alpha \mathcal{L}^{cont} =\frac{1}{N} \sum_{i} \mathcal{L}_{i}^{ce}+\alpha \frac{1}{N} \sum_{i} \mathcal{L}_{i}^{cont}.
\end{equation}

\section{Experiments Details}
To validate the performance of our model, we evaluate models with FCCL on three publicly available medical imaging datasets, namely EyePACS, HAM10000, and WSSS4LUAD. 
In addition, we compare our methods with some most commonly used UDA algorithms, namely DANN~\cite{ref_article4}, CDAN~\cite{ref_proc13}, and CDAN+E~\cite{ref_proc13}. 
The description of the datasets and implementation details are as follows.

\subsection{Datasets}
\textbf{The EyePACS dataset} is from Kaggle's Diabetic Retinopathy Detection Challenge~\cite{ref_url1}. 
The original labels are 0-4, indicating five different grades of diabetic retinopathy. 
According to ~\cite{ref_proc10}, the images are from cameras of different brands. Therefore, we follow their work to use the brand information as domain indicators. The dataset distribution is shown in Table~\ref{EyePACSdata}. We indicate that the division of training and testing data is according to the original challenge rules. 

\textbf{The HAM10000 dataset} is a multi-source dermatoscopic images dataset on kaggle~\cite{ref_article12}. 
The original labels characterize which of seven specific skin lesions are present in an image. 
The seven skin lesions are: (Nv), ..., and Df.
The dataset also provide age and sex information of each image. Here we use sex to divide the dataset into source (male) and target (female) domain. 
The dataset distribution is shown in Table~\ref{HAM10000data}.

\textbf{The WSSS4LUAD dataset} is a collection of image patches from whole slide tissue images of lung adenocarcinoma~\cite{ref_article13}. 
We simplified the original challenge task to a binary classification distinguishing whether tumor stroma is present in an image or not. 
The images have two major sources: Guangdong Provincial People's Hospital (GDPH) and The Cancer Genome Atlas (TCGA)~\cite{ref_url2}. 
We treat the two sources as our domain indicator in this task, namely GDPH(source domain) and TCGA(target domain).
The dataset distribution is shown in Table~\ref{WSSS4LUADdata}.

\begin{table}
\centering
\caption{Distribution of EyePACS dataset(train+val/test).}\label{EyePACSdata}
\begin{tabular}{p{70 pt}p{60 pt}p{50 pt}p{50 pt}p{50 pt}p{40 pt}}
\toprule
Camera Brand &  Grade 0 & Grade 1 & Grade 2 & Grade 3 & Grade 4\\
\midrule   
A(Source) &  16296/24784 & 1446/2306 & 3643/5358 & 566/748 & 511/858\\
B &  2862/4596 & 317/518 & 501/768 & 67/103 & 48/86\\
C &  2568/3808 & 225/324 & 512/726 & 166/193 & 69/133\\
D &  2978/4704 & 362/563 & 407/711 & 52/80 & 59/71\\
E &  627/980 & 62/78 & 141/159 & 12/23 & 12/24\\
\bottomrule
\end{tabular}
\end{table}


\begin{table}
\centering
\caption{Distribution of HAM10000 dataset(train+val/test).}\label{HAM10000data}
\begin{tabular}{p{60 pt}p{40 pt}p{40 pt}p{40 pt}p{40 pt}p{40 pt}p{30 pt}p{20 pt}}
\toprule
Sex & Nv & Mel & Bkl & Bcc & Akiec & Vase & Df\\
\midrule
Male(Source)   & 2714/707 & 556/133 & 504/122 & 254/63 & 182/39 & 60/9 & 55/8 \\
Female & 2589/648 & 372/91 & 340/84 & 160/37 & 74/32 & 60/13 & 46/6 \\
\bottomrule
\end{tabular}
\end{table}
\begin{table}
\centering
\caption{Distribution of WSSS4LUAD dataset(train+val/test).}\label{WSSS4LUADdata}
\begin{tabular}{p{80 pt}p{80 pt}p{80 pt}}
\toprule
Tissue Source  & Tumor Stroma & Non Tumor Stroma\\
\midrule
GDPH(Source)  & 4489/1146 & 2155/515\\
TCGA  & 1159/280 & 270/77\\
\bottomrule
\end{tabular}
\end{table}

\subsection{Implementation Details}
We used Resnet50 as the network backbone. We trained the network with stochastic gradient decent optimizer and set the learning rate schedule as half-cycle cosine with initial learning rate as 0.001 and one-epoch warm-up. The training last 200 epochs. The batch size was 32. The temperature coefficient $\tau$ was set to 1. For data augmentation, we applied random cropping of 512 × 512 pixels, random horizontal and vertical flip, random color and grayscale transformation, random gaussian blur, and random rotation. Our experiments are performed on 1 NVIDIA A100 GPU, and one experiment takes 24 hours with about 20G of the memory footprint. Our code will be open sourced after publication.

For a fair comparison with UDA methods, we further fine-tuned FCCL models with target domain images and auto-generated pseudo-labels. We use the class of model outputs with the highest confidence as the sample pseudo label. When the source domain is trained, $\alpha$ is set to 1, when the target domain is finetuned, $\alpha$ is set to 0, and the initial value of smoothing coefficient $m_0$ is set to 0.999. During finetune, the learning rate is set to 1e-6, and the training is performed for 1 epoch.

\section{Results}
We verified that in all the three datasets mentioned above, domain shifts exist and can affect classifier performance. We also verified that our proposed FCCL can effectively mitigate the domain shift in terms of both performance and feature distribution. 

\subsection{Domain Shift Affects Performance in Three Different Modalities} \label{Results1}
Table~\ref{domainShift} summarizes the performance drop caused by domain shift in the three datasets. 
For the EyePACS dataset, the performance numbers are in quadratic weighted kappa. The source domain refers to fundus camera manufacturer A. The target domain performance is averaged by all target domains (fundus camera manufacturers B, C, D, and E). Individual target domain performance numbers are in our supplementary materials. 
For the HAM10000 dataset, the performance numbers are in AUC for predicting each class, averaged over all the seven classes. The source domain refers to male data and the target domain refers to female data. 
For the WSSS4LUAD dataset, the performance number is in AUC for predicting stroma vs. non-stroma tissue. The source domain refers to GDPH data and the target domain refers to TCGA data. 
For all the three datasets, there is clearly a performance drop in the target domain. 

\begin{table}
\centering
\caption{Quantitative measure of domain-shift-caused performance drop.}\label{domainShift}
\begin{tabular}{p{70 pt}>{\centering}p{80 pt}>{\centering}p{80 pt}>{\centering\arraybackslash}p{70 pt}}
\toprule
Domain & EyePACS & HAM10000 & WSSS4LUAD\\
\midrule
Source & 0.799 & 0.974 & 0.982\\
Target & 0.734\textcolor[rgb]{0,0.5,0}{(-6.5\%)} & 0.926\textcolor[rgb]{0,0.5,0}{(-4.8\%)} & 0.946\textcolor[rgb]{0,0.5,0}{(-3.6\%)}\\
\bottomrule
\end{tabular}
\end{table}

\subsection{FCCL Improves Model Performance in Target Domain}
In Table~\ref{solveDomainShift} we list the target domain performance after applying FCCL 
as well as some other UDA techniques. The target domain performance is computed in the same manner as in ~\ref{Results1}. Compared with Table~\ref{domainShift}, we see that our proposed FCCL method is effective in improving the target domain performance for all the three datasets. Though other UDA techniques are also effective, FCCL exceeds their performance without seeing any target domain images. If unlabeled target domain images are available, fine-tuning can further improve the performance as the last row in Table~\ref{solveDomainShift} suggests. 

\begin{table}
\centering
\caption{Target domain performance of FCCL and different UDA techniques.}\label{solveDomainShift}
\begin{tabular}{p{100 pt}>{\centering}p{70 pt}>{\centering}p{70 pt}>{\centering\arraybackslash}p{70 pt}}
\toprule
Method & EyePACS & HAM10000 & WSSS4LUAD\\
\midrule
DANN & 0.735 & 0.936 & 0.941\\
CDAN & 0.748 & 0.949 & 0.945\\
CDAN+E & 0.746 & 0.940 & 0.942\\
\hline
FCCL & 0.767 & \textbf{0.950} & 0.949\\
FCCL w/ Fine-Tuning & \textbf{0.788} & 0.948 & \textbf{0.950}\\
\bottomrule
\end{tabular}
\end{table}

\subsection{FCCL Reduces Domain Shift and Increases Class Separation}
\begin{figure}
\includegraphics[width=\textwidth]{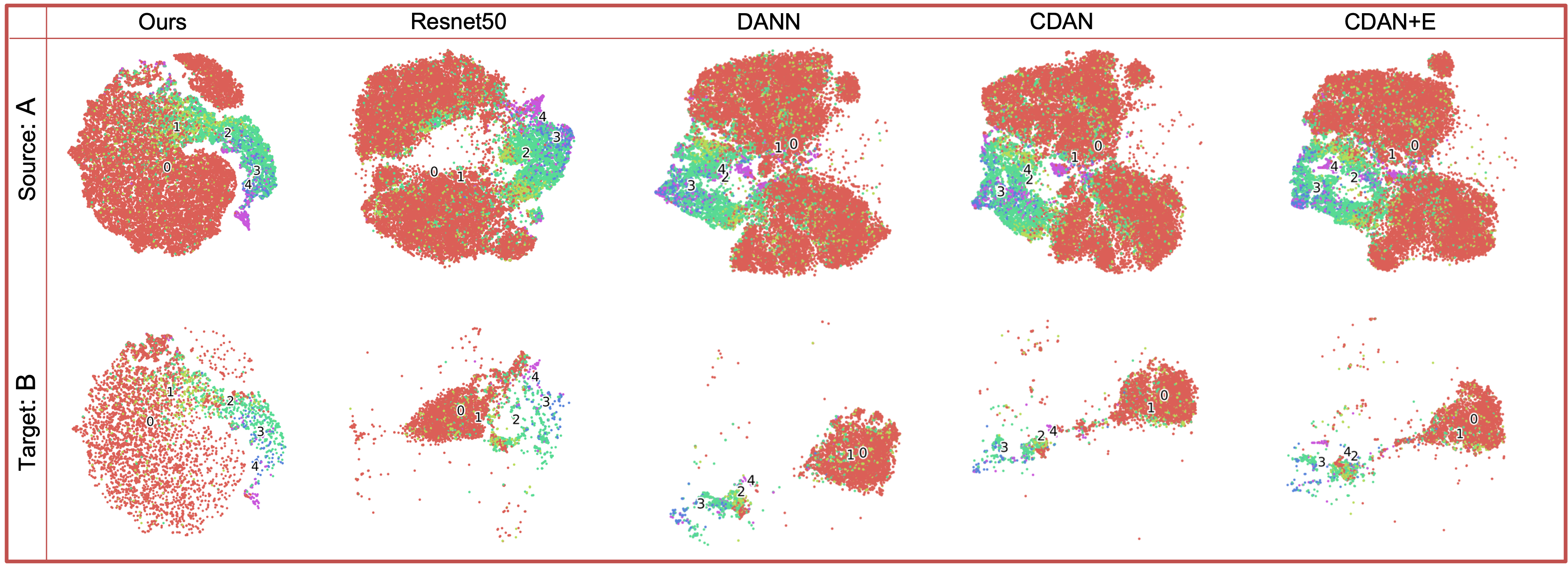}
\caption{Domain shift visualization with T-SNE on diabetic retinopathy classification task. Different colors represent different diabetic retinopathy grades. Top row: source domain features of different methods. Bottom row: target domain features.} \label{TSNE}
\end{figure}

We visualize the feature space distribution through various methods. Fig.~\ref{TSNE} is a T-SNE visualization on the features of the EyePACS image data. For FCCL generated features, the target domain features share roughly the same range as of the source domain. On the contrary, for features generated by other methods, there are clear gaps between source and target domains, suggesting the domain shift problem still exists on the feature level.

\begin{figure}
\includegraphics[width=\textwidth]{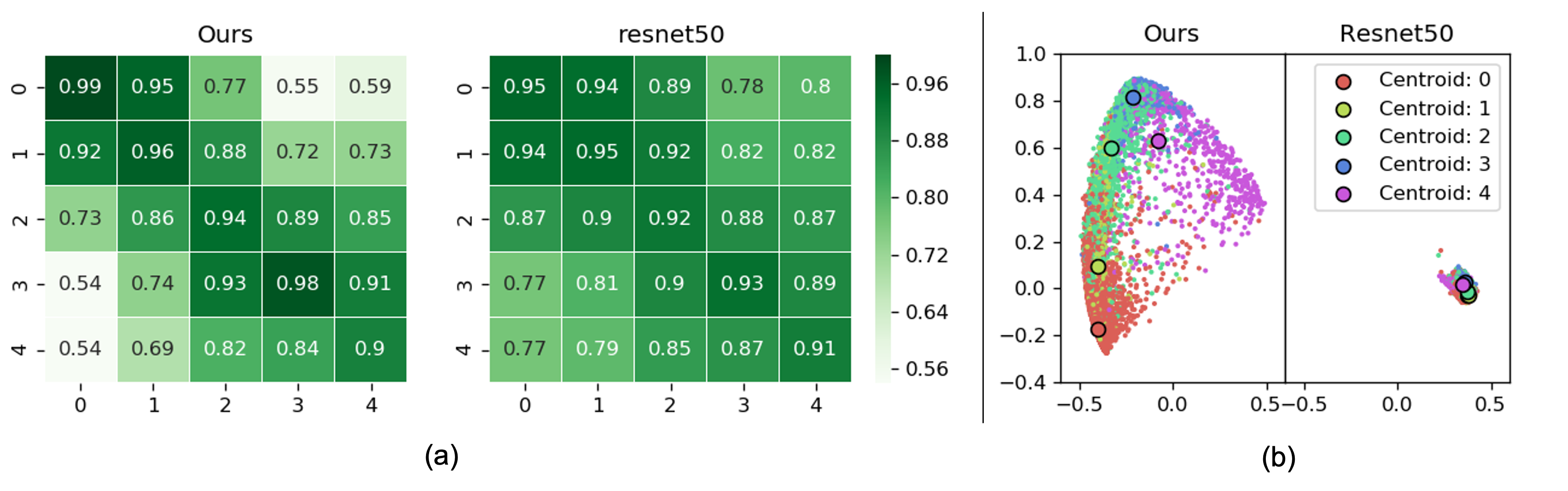}
\caption{Visualization on the spatial distribution of features in diabetic retinopathy classification task. (a) Heatmaps representing the degree of aggregation. The numbers in each grid are the mean of the cosine distances of the centroids and all sample features of each class. The main diagonal line indicates that the intra-class is more aggregated, and the non-main diagonal line indicates that the inter-class is farther away. (b) First two dimensions after reducing the high-dimensional features with PCA. FCCL leads to larger feature distribution space.} \label{hotmap}
\end{figure}

Fig.~\ref{hotmap}(a) are the heatmaps showing mean cosine similarity values of the features belonging to each class with respect to the centroids of all classes. The main diagonal terms of FCCL heatmap are generally larger than that of Resnet50, indicating that FCCL makes the features more clustered within the class. The off-diagonal terms of FCCL are generally smaller than that of Resnet50, indicating that FCCL keeps different classes of features away from each other. As can be seen in Fig.~\ref{hotmap}(b), FCCL makes the features more widely distributed in space compared with ResNet50. This is beneficial to reducing the effect of domain shift.

\section{Discussions}




In this paper, we validate multiple types of domain shift problems for medical imaging datasets.
The first is to divide domains by medical imaging equipment manufacturer. For the fundus photography dataset, we use camera brands to distinguish source and target domain, which is a more "pure" definition. For this task with a clear domain, the domain adaptation method can significantly reduce the domain shift and improve performance.
The second is to divide the domain according to the target population. For the dermoscopic images dataset, we divide it into two domains according to sex. The results show that the domain shift of males and females exist, but may not be as large as imaging equipment brands. In the future, we will explore other demographic aspects such as age, race, etc. to further verify domain shift of population.
The third is to divide domains by different data sources. In the H\&E stained tissue images dataset, GDPH data represent a single hospital as data source, with the same imaging equipment and tissue processing protocol. However, TCGA data are from a combination of multiple sources (hospitals). We treated TCGA as one target domain, which may degrade other UDA performance and make an unfair comparison. We notice that in principle, FCCL still works in the settings where the target domain definition is not "pure". 

Our method makes the intra-class features more clustered and the inter-class features farther, which can reduce the domain shift. At the same time, after the target domain is fine-tuned, some space can be sacrificed in exchange for further improvement in the performance of the target domain. This is mainly due to two adjustments made during fine-tune, one is to add the mean and variance of the target domain to the Batch Normalization(BN) layer, and the other is to adjust the model weights and biases. In the future work, we will further verify the influence degree of these two adjustments.

\section{Conclusions}
Our proposed FCCL can improve medical image classification model performance when domain shift problem exists. FCCL is effective for at least three medical imaging modalities, i.e., fundus photography, dermoscopic images, and H\&E stained tissue images. Also, FCCL can work with labeled image data from only a single source domain and no target domain, making it useful under extreme data constraint.



\end{document}